\documentclass[preprint,12pt]{elsarticle}
\usepackage{hyperref}
\usepackage{enumitem}
\usepackage{amsmath}
\usepackage{tabularx}
\usepackage{setspace}

\usepackage{amssymb}

\journal{Information Fusion}

\begin{document}

\begin{frontmatter}



\title{Expanding Frozen Vision-Language Models without Retraining: Towards Improved Robot Perception
}


\author[inst1]{Riley Tavassoli}

\affiliation[inst1]{organization={San Diego State University},
            addressline={5500 Campanile Dr}, 
            city={San Diego},
            postcode={92182}, 
            state={California},
            country={United States of America}}

\author[inst1]{Mani Amani}
\author[inst1]{Reza Akhavian\corref{cor1}}
\ead{rakhavian@sdsu.edu}
\cortext[cor1]{Corresponding author}

\begin{abstract}
Vision-language models (VLMs) have shown powerful capabilities in visual question answering and reasoning tasks by combining visual representations with the abstract skill set large language models (LLMs) learn during pretraining. Vision, while the most popular modality to augment LLMs with, is only one representation of a scene. In human-robot interaction scenarios, robot perception requires accurate scene understanding by the robot. In this paper, we define and demonstrate a method of aligning the embedding spaces of different modalities (in this case, inertial measurement unit (IMU) data) to the vision embedding space through a combination of supervised and contrastive training, enabling the VLM to understand and reason about these additional modalities without retraining. We opt to give the model IMU embeddings directly over using a separate human activity recognition model that feeds directly into the prompt to allow for any nonlinear interactions between the query, image, and IMU signal that would be lost by mapping the IMU data to a discrete activity label. Further, we demonstrate our methodology's efficacy through experiments involving human activity recognition using IMU data and visual inputs. Our results show that using multiple modalities as input improves the VLM's scene understanding and enhances its overall performance in various tasks, thus paving the way for more versatile and capable language models in multi-modal contexts.
\end{abstract}




\end{frontmatter}


\section{Introduction}
\label{sec:sample1}
Multi-modal research in vision, audio and language has gained traction in recent years\cite{UPPAL2022149, NGUYEN2023101868}, and now with current studies showing that Large language models (LLMs) have the capabilities of complex question answering and reasoning \cite{huang2023reasoning}, there has been an influx of attention towards utilizing multi-modal LLMs.  Recent research on vision-language models has further shown these reasoning capabilities can be extended to other modalities \cite{liu2023prismer}. In this paper, we propose a method that extends frozen, pretrained visual-language models to understand inertial measurement unit (IMU) data while being extensible to any other modality.
This method of extending pretrained models without retraining or finetuning reduces training costs dramatically in an era of deep learning where it has become infeasible to train most models from scratch for the majority of researchers and developers \cite{Wang_2020_Expensive}. At these large sizes, models can learn abstract, generalizable reasoning skills that are difficult to replicate in smaller models \cite{bubeck2023sparks}. Specifically, language models present a new paradigm of foundation models that offer unlimited downstream use cases, with the limitation of text being the singular modality. Vision-language models (VLMs) have allowed for images to be interwoven with text, taking advantage of the skills the base LLM learned while being trained on text. Flamingo \cite{alayrac2022flamingo} proposed a novel VLM architecture where trainable layers were injected into the frozen LLM. These new trainable layers require far less training than the base LLM while allowing the raw image embeddings to be processed in a depth-wise manner alongside the accompanying text. This results in the frozen LLM being capable of understanding image embeddings that have been resampled to match the distribution the LLM expects. This allows for the LLM's in-context learning capabilities to be used on images, making the model versatile and removing the need for fine-tuning to a domain-specific dataset \cite{dipalo2023unified}. Instead of training new layers or modules for every additional modality to be incorporated, any modality can arbitrarily be aligned to share the embedding space of the vision encoder through contrastive training. Consequently, the layers that translate vision embeddings into representations the LLM understands also work on any other modality that has been aligned with the vision embedding space.

\begin{figure}
    \centering
    \includegraphics[width=0.9\linewidth]{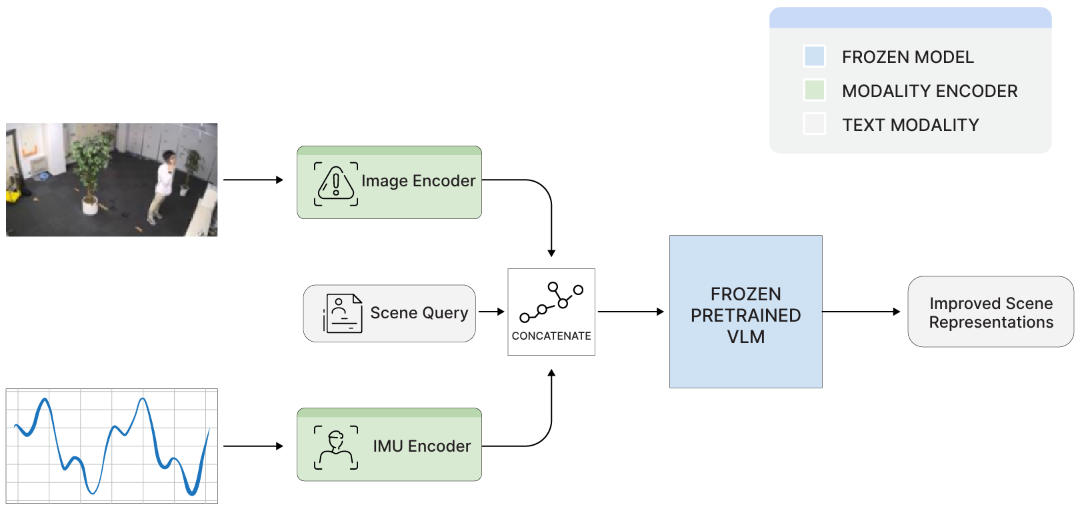}
    \caption{Overview of the approach showing the concatenation of multiple modal representations of a scene with a query yielding better, more semantic responses}
    \label{fig:OverallIdea}
\end{figure}

Most contrastive learning methods rely on large datasets, but with the methods we propose in this paper, even modalities with relatively few examples can sufficiently align their embedding space to the vision encoder. This idea also addresses a growing demand for larger generalist models to use any modality to enable users to take advantage of the abstract representations it has learned. As such, our main contributions in this paper are as follows:
\begin{enumerate}[nosep]
    \item A methodology that allows for the extension of frozen, pretrained vision-language models to accommodate any number of modalities, broadening their applicability and versatility.
    \item An understanding of how multi-modal inputs contribute to the development of increasingly nuanced scene representations, adding depth and context to machine understanding.
    \item A validated evidence that the integration of various modalities improves scene understanding, a critical aspect of machine perception and cognition.
    \item A demonstration of how relatively small datasets can be used for contrastive alignment.
\end{enumerate}

\subsection{Robot Perception and Human Activity Recognition (HAR)}
The goal of this paper is to leverage VLMs for better scene understanding toward improved robotics perception, especially in human-robot interaction (HRI) scenarios. In this regard, human activity recognition (HAR) using wearable devices can help robots better perceive their environment by gaining an understanding of the type of activity in which the human counterpart is engaged with. Because there already exist very competent HAR models, we choose to supply the IMU embeddings directly in the prompt to assess model performance on more granular aspects of scene understanding that are not readily extractable with pre-existing models. For practical and automated HRI applications, the HAR classification could also be retrieved from an auxiliary model and appended to the query.


\section{Related Work}
HRI has garnered interest in recent years manufacturing due to the potential production efficiency improvement. However, robots do not have the innate contextual scene understanding capability humans latently possess \cite{RobotPerception}. To remedy this issue, researchers have conducted extensive research into both robot perception \cite{RobotPerception2} and robot action \cite{RobotAction}

For robotic action,  RT-2 \cite{rt22023arxiv} is a visual-language-action (VLA) Model that leverages visual-language models (VLM) such as PaLI-X \cite{chen2023palix} to generate action commands for a robot based on a query. The model changes the backbone VLM's generated token to include robotic actions that are used as inputs to the low-level robotic controller. Principally, the paper shows the potential of adapting VLMs to VLAs by combining VLM pretraining with robotic data.
\par

PaLM-E \cite{Driess2023PaLMEAE} is an embodied, multi-modal language model capable of executing complex robotic tasks via planning and scene understanding. However, they use a training paradigm different from the one presented in this paper whereby modality encoders are trained end-to-end with a frozen LLM as the head of the encoder, similar to\cite{Frozen}. Importantly, they highlight the ability of LLMs to reason on multi-modal inputs in a way similar to language. There are several other recent works, such as ImageBind \cite{girdhar2023imagebind},  that integrate multiple modalities into a unified embedding space, showing how a linear combination of embeddings from multiple modalities yields a better representation. These developments highlight the capabilities of multi-modal learning and underscore the importance of exploring it further. Macaw-LLM \cite{lyu2023macawllm} provides a new architecture for multi-modal inputs, published as a vision-audio-text LLM with the introduction of a new architecture containing an alignment module that aligns the embeddings of multiple modalities to a common embedding space. The benefit of what we design in this work is its ability to leverage pretrained models without the need for a new architecture or retraining of the base model or an alignment module. Works such as BlIP-2 \cite{li2023blip2} follow the same philosophy of feeding different modalities to language models in a format that they can understand and process through a specific pretraining regime. BLIP-2 combines "off the shelf" frozen image encoders and frozen LLMs and proposes a pretraining strategy that bridges the gap between modalities. \cite{li2023blip2} show a novel architecture for a light-weight HAR model designed for processing videos and trained contrastively on associated activity captions.VLMs have been employed in the past for HAR. One such instance is VicTR, a model that utilizes a joint video and text encoder to create video-conditioned text tokens for improved HAR \cite{kahatapitiya2023victr}. In another study, the authors developed a HAR algorithm utilizing wide time-domain convolutional neural network and multi-environment sensor data for daily behavior recognition while using contribution significance analysis to assess the contribution of each sensor to the detection of the activity \cite{HAR10}.
\par
PaLM-E’s approach to integrating sensor signals as inputs in a multi-modal language model provided valuable insights into the potential capabilities of LLMs to reason on multi-modal inputs in a similar manner to language. However, they rely on a paradigm that requires the encoders to be trained end-to-end with a frozen LLM, limiting the flexibility of the system. ImageBind \cite{girdhar2023imagebind} integrates multiple modalities into a unified embedding space through contrastive alignment, bypassing the high training cost of using the LLM directly. Our work strives to develop a methodology that allows LLMs to accommodate an arbitrary number of modalities without needing a new architecture, an issue faced by works like Macaw-LLM \cite{lyu2023macawllm}.


\vspace{10pt}
\section{Methodology}\label{Sec:Introduction}
\begin{figure}
    \centering
    \includegraphics[width=1\linewidth]{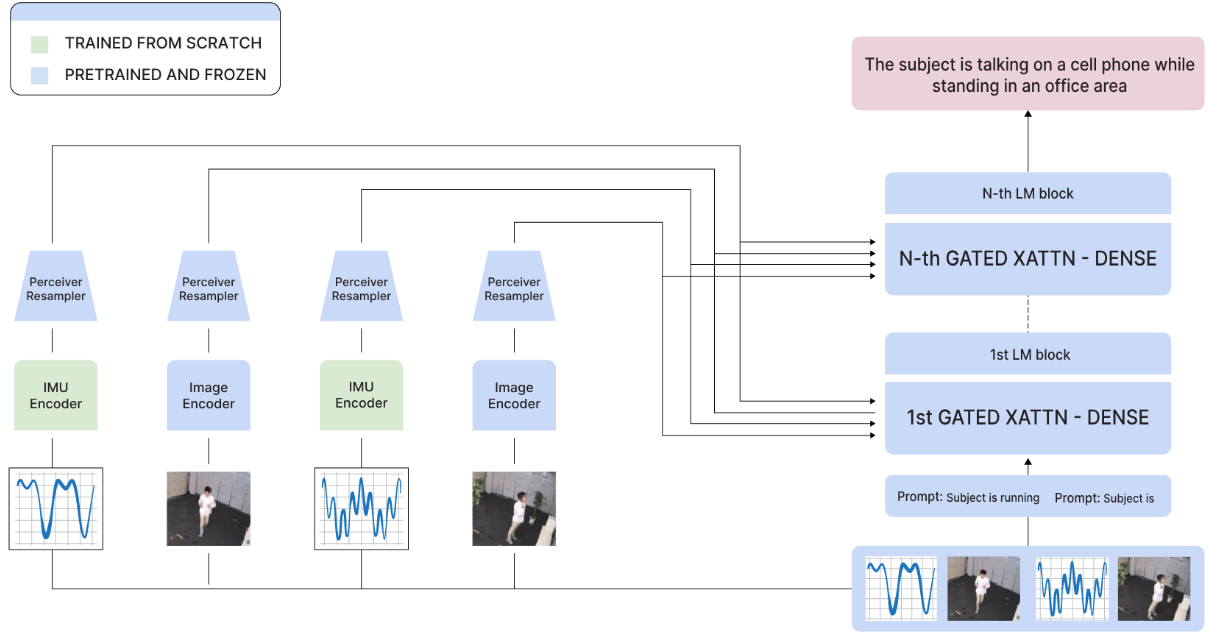}
    \caption{The architecture of Flamingo VLMs extended to handle image-IMU pairs of inputs}
    \label{fig:Perceiver}
\end{figure}

Figure \ref{fig:Perceiver} shows how raw inputs are processed through the VLM. An important distinction is that we linearly combine the image and IMU embeddings for a single example after having passed through the perceiver resampler but before they pass through the gated cross attention layers. This linear combination of encoded representations provides the VLM with a more holistic representation of the scene. The training method that aligns the IMU encoder to the pretrained visual encoder is outlined below.

\subsection{Dataset}

We use the MMAct dataset \cite{Kong_2019_ICCV} which consists of 35 different human actions with varied durations of ~2-10 seconds. Each sample is recorded on 4 different cameras at different points in the room. The room is also set up in 4 distinct ways with different obstacles or components. Example actions include talking on the phone, running, and pulling a cart. We concatenate all IMU signals, down sampling where necessary such that each signal is sampled at 50 Hz. Each signal provides 3 channels, and with 4 signals (two accelerometers at different parts of the body, a gyroscope, and a magnetometer), we attain a 12 channel IMU signal. We sample a window of length 256 from this data, padding with zeros when the signal is not long enough. We randomly sample a frame from the corresponding video. We use a batch size of 512 and train for a maximum of 100 epochs, early stopping once the validation loss is minimized. The total train size is 6093 examples.

\subsection{Modality Encoders}

 In this work, we extend visual language models to understand IMU data encoded using a transformer-based encoder in combination with a 1-d convolution without retraining the visual language model. To train this encoder, we contrastively optimize temporally overlapping image-IMU pairs to have a large cosine similarity using a frozen, pretrained ViT-L/14 \cite{Radford2021LearningTV} as the visual encoder. An extension of CLIP for video, X-CLIP \cite{xclip}, has previously been explored by the authors for HAR \cite{shahavaz2023robust}. The presented work seeks to show the capability of extending VLMs understanding to multiple modalities with no retraining. Therefore, we are constrained to the frozen visual encoder the VLM was trained with. This is because as we contrastively train our IMU encoder to share the embedding space of the vision encoder, it is necessary that this shared embedding space towards which the IMU encoder optimizes is the same as the embedding space the VLM was trained to understand. Had we used a different vision encoder to contrastively train the IMU encoder, the pretrained VLM would not understand the IMU embeddings without retraining. Here, we are inspired by the work presented in ImageBind \cite{girdhar2023imagebind} to train arbitrary modality encoders to align their embeddings with a desired embedding space.



\subsection{Contrastive Pretraining}
Contrastive learning, a subfield of unsupervised learning, works by learning a representation of its inputs such that similar inputs result in similar vectors and dissimilar inputs yield dissimilar vectors \cite{ContrastiveReview}. It has been successfully applied in a variety of machine learning tasks, ranging from image and speech recognition to natural language understanding, largely due to its effectiveness in learning rich, meaningful embeddings from unlabeled data. Multi-modal contrastive learning is still an active area of research \cite{contrastiveresearcg, nakada2023understanding} where the loss functions with which we optimize over are just beginning to be explored. When contrastively training an encoder for a modality with a temporal dimension such as IMU data, the window size is directly correlated with information content which makes it an important hyperparameter to tune and optimize for good representation quality \cite{contrastive}. 
We utilize a symmetric cross entropy loss objective, also known as the infoNCE loss \cite{originalinfopaper, INFOpaper}, in order to train our IMU encoder model. The loss maximizes the dot product of matching pairs in a batch and minimizes the dot product of negative pairs. This was most recently popularized in a multi-modal context with CLIP  \cite{Radford2021LearningTV}.

\begin{equation}
   L_{\text{{infoNCE}}} = - \sum_{(i, j) \in P} \log \left( \frac{e^{\text{{CoSim}}(z_{i}, z_{j}) / \tau}}{\sum_{k=1}^{N} e^{\text{{CoSim}}(z_{i}, z_{k}) / \tau}} \right)
\end{equation}

For every pair (i,j) in set P, which represents positive pairs of data instances, we compute the cosine similarity CoSim between the respective representations \begin{math}z_{i}\end{math} and \begin{math}z_{j}\end{math}. This similarity score is scaled by a temperature parameter \begin{math} \tau \end{math} to control the sharpness of the distribution. The logarithm of this ratio is then computed, and the loss is the negative sum over all positive pairs. This formulation encourages the network to maximize similarity for positive pairs and minimize similarity for negative pairs, where positive pairs are defined as images and overlapping IMU windows, and negative pairs are images and non-overlapping IMU windows.
We also add a supervised loss term to the loss function, mapping the embedded IMU representation to class logits with a linear head. This enforces a constraint on the embedding space that keeps embedded actions linearly separable. With the addition of this supervised loss term, we observed more specific, distinct outputs from the VLM when given IMU embeddings.
\par
Rather than computing the infoNCE and supervised losses on the outputs from the encoders, we further process both encoded representations by passing them through the frozen, pretrained perceiver resampler module. This outputs a predefined set of latent vectors that are resampled representations of the input. For our implementation, we map an encoded sequence length of 256 with dimension 1024 to a length of 64 with the perceiver resampler. We then average pool along the sequence dimension for both image and IMU embeddings to obtain 1-d vectors of size 1024 for each sample. It is with these representations we compute the infoNCE and supervised loss terms. In our empirical tests, this process of including the perceiver resampler module grounds the representation the IMU encoder learns more rigidly. We observed this in testing different iterations on an activity recognition sub-task where we prompt the VLM with only IMU embeddings to identify the action being performed. IMU encoders trained without the perceiver resampler exhibited far worse performance on this task, such that when combining the IMU embeddings with vision embeddings, worse performance could sometimes be observed. Our hypothesis for why we see better performance with this architecture is that the inclusion of the perceiver resampler strictly constrains the features learned by the IMU encoder to have a similar distribution to the features of the image encoder. When computing loss on the embeddings that are output from the encoders rather than the perceiver resampler, the loss is far noisier whereas the perceiver resampler processes embeddings of both modalities into a shared distribution.
This contrastive and supervised learning objective enables the IMU encoder to learn a meaningful mapping from raw sensor data to a representative embedding space that aligns with the image encoder. Most unsupervised methods, contrastive learning included, require large amounts of data. This paper explores how a relatively small training dataset of around 6,000 image-IMU pairs can be leveraged to align an IMU encoder with a vision encoder.

\subsection{ Multi-Modal Large Language Model}
We utilize VLMs as a high-level reasoning module to better understand a scene given various modal representations. We use an implementation of the Otter VLM \cite{li2023otter}, a continuation of Open Flamingo \cite{openflamingo}, the open-sourced version of the original DeepMind paper, Flamingo \cite{alayrac2022flamingo}. Otter is further trained on an instruction dataset to support multi-modal in-context instruction tuning which involves conditioning the language model on the corresponding media, such as an image, that corresponds to a caption or an instruction-response pair \cite{li2023otter,li2023mimicit}. This makes Otter deliver more guided outputs when we effectively design prompts.
\par
VLMs such as Otter are autoregressive decoder-only models, with image embeddings represented in the tokenizer with a special token. The image embeddings, or any other modality’s embeddings, are passed through a module introduced in the original Flamingo paper called the Perceiver Resampler which takes as input the sequence output of a transformer encoder and outputs a fixed set of resampled tokens as based on a set of learnable latent queries. This allows for variably sized inputs to be mapped to the same length of tokens, and it allows the frozen, pretrained language model to resample the static image embeddings throughout the layers of the LLM. 

Because Otter was trained on an instruction-tuning dataset, the model learns to structure its response to follow the preceding in-context examples which allows us to query the model’s understanding of the scene. In this paper, we show that the addition of the IMU embeddings in the prompt allows the VLM to better reason about a scene and more wholly understand the activities of the humans present in the visual input. 



\section{Experiments}
\begin{figure}[h]
    \centering
    \includegraphics[width=1\linewidth]{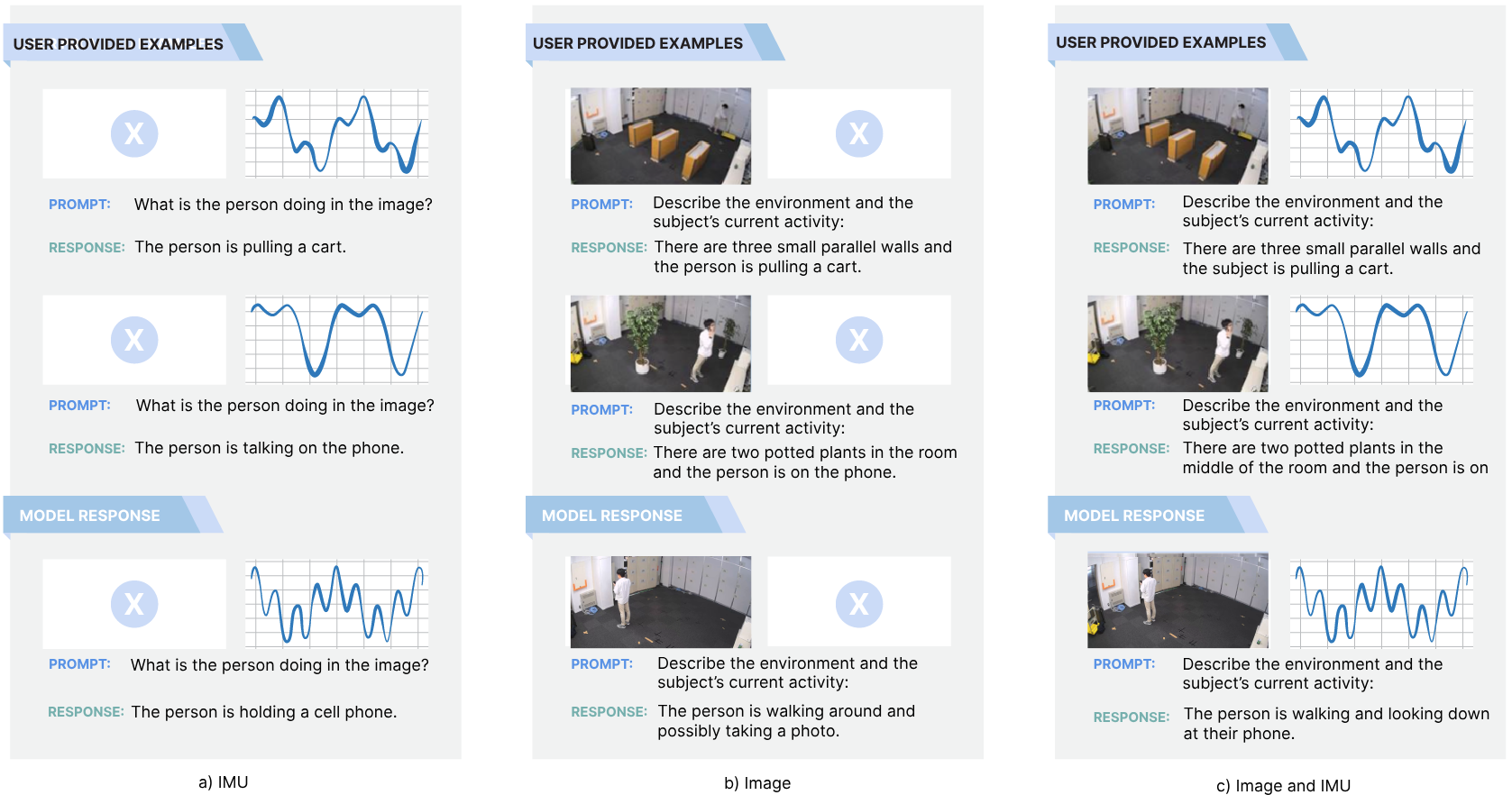}
    \caption{In-context generation with only IMU, only images, and both modalities}
    \label{fig:ChatExamples}
\end{figure}

Below, we show the capabilities of the pretrained Otter model on semantic scene understanding tasks when provided vision data, IMU data, and a combination of both. We take advantage of conditional generation by prepending our query with two example question-response pairs to update the model’s output distribution to be more in line with our expectations. Figure \ref{fig:ChatExamples} shows model responses given different combinations of input modalities.

\section{Results}

\begin{figure}
    \centering
    \includegraphics[width=1\linewidth]{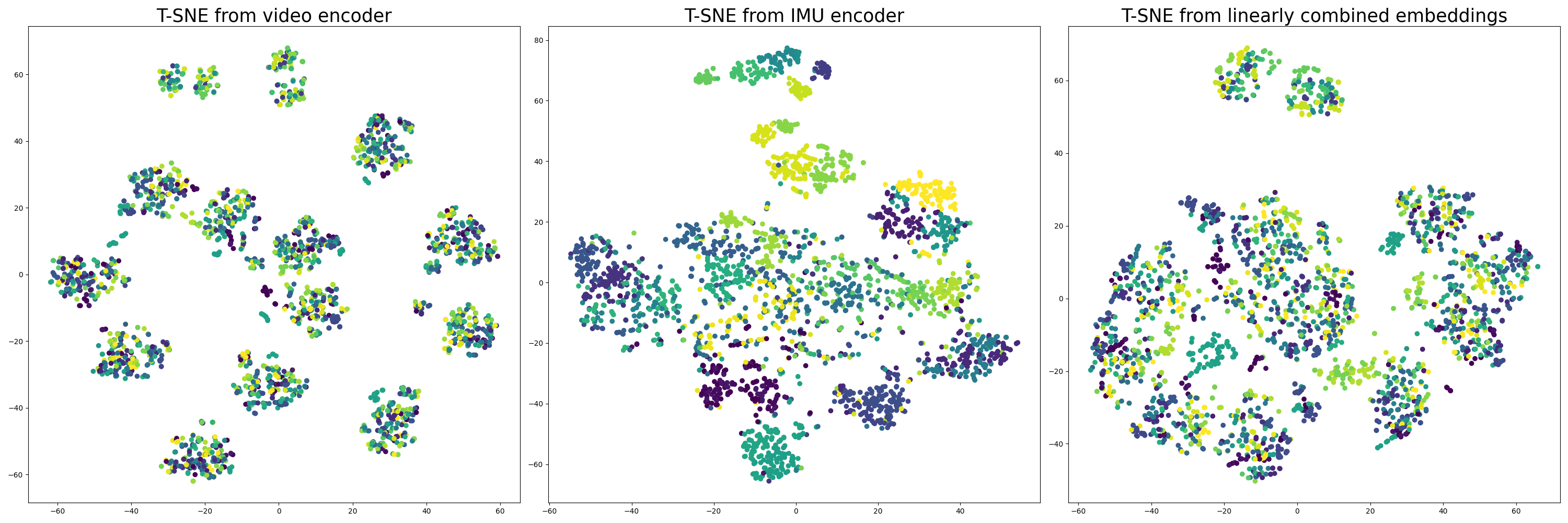}
    \caption{t-SNE visualization of video and IMU encoder embeddings across 35 classes}
    \label{fig:T-SNE}
\end{figure}

We evaluated the effectiveness of our contrastive alignment process by mapping the embeddings of the IMU and image data for each of the 35 classes via t-distributed stochastic neighbor embedding (t-SNE), a technique used to visualize high-dimensional data in a way that shows underlying data distributions in a 2-dimensional representation that can easily be plotted \cite{t-SNE}. Figure \ref{fig:T-SNE} shows the result of this visualization where each class is represented by a different color, and the clusters suggest distinctive patterns in the data.

The video encoder embeddings display clear clusters, suggesting that the model successfully extracted meaningful patterns from the data. However, these clusters do not align with the specific activities performed as there are no clear color groupings, an outcome we anticipated. This is because the image encoder was not specifically fine-tuned for HAR.
The IMU encoder embeddings lack some of the structure present in the image embeddings, suggesting that the contrastive alignment of encoders did not fully align the two models to share the exact same embedding space, but the class distribution is far more organized which allows the model to better understand a user's actions as evidenced the very clear color groups. Further, the two modalities are fundamentally capturing different characteristics of the scene, which is by design, but that does mean that the embedding space of the IMU encoder will naturally have a different structure even after alignment. For example, the IMU data more closely corresponds to what a person is doing, e.g. two different people doing the same activity have more similar IMU data than images of two people doing the same activity due to the potential for different backgrounds, environments, or peripheral objects. Because the IMU data contains less total information than the associated images, there will be some breakdown in structure where the image embeddings more finely correspond to a given input.

We also test how linearly combining the embeddings from both modalities changes the shared embedding space when visualized with t-SNE. In our experiments, we see that the weights used in linearly combining the two modal embeddings interpolate between the structure of the video and IMU embedding spaces. For Figure \ref{fig:T-SNE}, we weight the vision embeddings 80\% and the IMU embeddings 20\%. In practice, these values must be empirically tuned to maintain the structure of the desired embedding space while gaining some smaller amount of information from the new embedding space. ImageBind \cite{girdhar2023imagebind} exploits the linear combination of vectors in multi-modal embeddings spaces for the task of video retrieval, obtaining better results when using audio in conjunction with vision data.

This emergent structure of grouped examples of the same activity that is present in the IMU embedding space and not in the vision embedding space indicates that the raw IMU distribution is more implicitly linked to the activity label. We view this as a feature allowing the two modalities to naturally complement one another, the IMU data encoding the kinematic, activity data the vision encoder struggles to encode. This point can be seen in Table \ref{tab:modalityComparison}. This table shows the linear probe performance of a supervised HAR model trained by video only, IMU only, and combined video-IMU data. The IMU embeddings naturally encode information about an individual’s action with far less noise than is present in an image. When combining modalities, we concatenate the output embeddings of each encoder, mapping the combined vector to class logits. This shows that the contrastive alignment of modalities can provide novel information to a pretrained model that otherwise would not be present in the unimodal data. This hypothesis warrants further investigation in future studies.

\begin{table}[ht]
\centering
\begin{tabular}{|l|m{2cm}|m{2cm}|m{2cm}|m{3cm}|}
\hline
\textbf{Modality} & \textbf{Training Loss} & \textbf{Test Loss} & \textbf{Training Accuracy (\%)} & \textbf{Test Accuracy (\%)} \\
\hline
Video & 1.0428 & 1.4748 & 65.07 & 52.41 \\
IMU & 0.4052 & 1.2468 & 91.83 & 64.47 \\
Combined & \textbf{0.2138} & \textbf{0.8753} & \textbf{94.58} & \textbf{74.46} \\
\hline
\end{tabular}
\caption{Supervised activity recognition for different modalities. Despite sharing the same embedding space, each modality still preserves unique information, as reflected in the increased performance when combining embeddings}
\label{tab:modalityComparison}
\end{table}

\section{Conclusion}
In this paper we have proposed a methodology to extend pretrained vision language models to any modality by only training the new encoder. We have shown the ability of a contrastive and supervised objective to sufficiently map an encoder to the pretrained vision encoder’s embedding space thereby taking advantage of the pretrained VLM. Further, we have shown how multiple modalities leads to a more robust and general scene representation and highlighted its potential in collaborative robotics.

\subsection{Future Work}
Future work can explore the effects of larger VLMs or VLMs with different architectures with multi-modal fine-tuning. The model size can prove as a limitation to the quality of the models responses, however, larger models will have longer inference times which could prove as an issue in different implementations. We plan to implement multi-modal fine-tuning to models such as MiniGPT-4 \cite{zhu2023minigpt} which uses a base Vicuna model \cite{vicuna2023} as the backbone LLM to assess and compare their capabilities with the Otter model. The MiniGPT-4 utilizes a Q-former with a frozen pretrained vision encoder compared to gated cross-attention layers which the flamingo model uses. 

\begin{figure}
    \centering
    \includegraphics[width=0.5\linewidth]{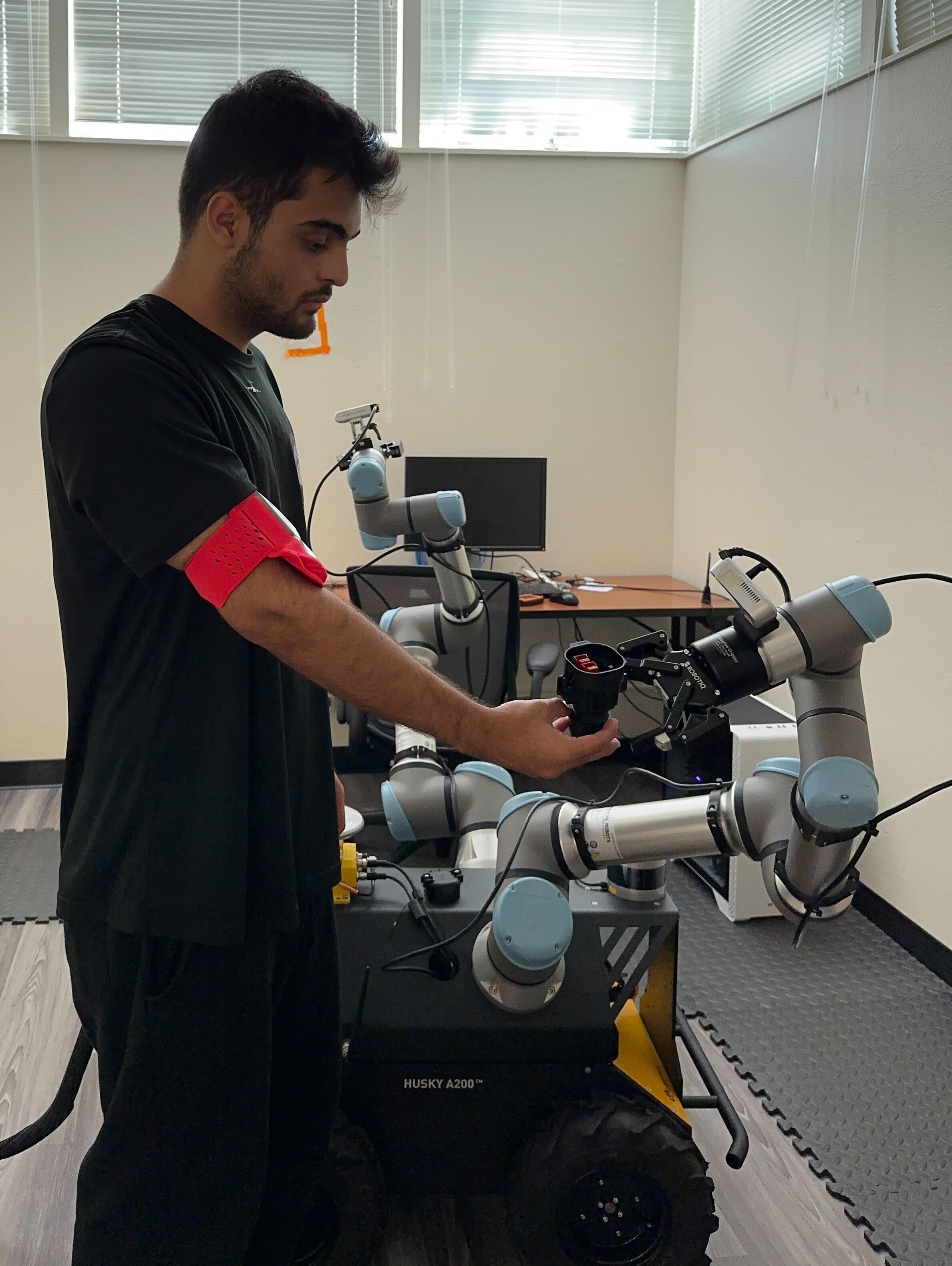}
    \caption{A researcher investigating multi-modal robot perception for human-robot collaboration}
    \label{fig:coworking}
\end{figure}

Another area we plan to explore is the assessment of information quality of each modality. We hypothesize that modalities can have varying levels of generalizable information regarding the activity and how we identify and account for these discrepancies. Previous work such as ImageBind uses multi-modal embedding space arithmetic to linearly combine embeddings of different modalities to yield a more information-dense embedding vector. \cite{girdhar2023imagebind}.
The outcome of the presented work will be ultimately used in the context of HRI for better robot perception. The authors are currently exploring HRI scenarios in the context of construction activities where visual data from robot cameras and IMU data from wearable sensors by construction workers are used to enhance robot perception as seen in Figure \ref{fig:coworking}.

Other avenues of future research can be implementing the methodology introduced in RT-2 which consists of utilizing robotic data through the VLM pretraining. We introduce the feasibility of extending VLMs to any number of modalities and experiment with the viability of implementing modality extensions on VLAs.
The study of contrastive alignment of modality encoders to a shared embedding space is an avenue we plan to explore with new training objectives and data-dependent significance analysis across multi-modal representations.
\subsection{Limitations}
The Otter model uses MPT-7b as its backbone, making it fast for inference, but with technical limitations in its hallucinations. Further, because the dataset of video frame-IMU pairs used, MMAct, is relatively small, we do not attain a 1:1 representation between video and IMU data. This is expected as IMU data intrinsically has a data distribution distinct from any correlated video frames.
Another downfall of extending modalities without pretraining is that the learnable model parameters have not been trained for multi-modal processing, potentially causing an increase in hallucinations. Current research indicates that poor training and low-quality training data has a direct effect on the degrees of hallucination \cite{Limitation1}.  
\section*{Conflict of Interest}
The authors declare that they do not identify any personal relationships or financial ties that would affect the contents or the publishing of this paper. 
\section*{Data Availability}
Source code and data will be made available upon request.
\section*{CRediT authorship 
contribution statement}
\textbf{Riley Tavassoli} Conceptualization, Methodology, Software, Investigation, Validation, Writing - Original Draft, Writing - Review \& Editing, Visualization.
\newline
\textbf{Mani Amani}: Investigation, Validation, Visualization, Writing - Original Draft, Writing - Review \& Editing, Software. 
\newline
\textbf{Reza Akhavian}: Project administration, Funding acquisition, Writing - Review \& Editing, Supervision.

\section* {Declaration of Funding}
The presented work has been supported by the U.S. National Science Foundation (NSF) CAREER Award through the grant \# CMMI 2047138. The authors gratefully acknowledge the support from the NSF. Any opinions, findings, conclusions, and recommendations expressed in this paper are those of the authors and do not necessarily represent those of the NSF.
 \bibliographystyle{elsarticle-num} 
 \bibliography{cas-refs}





\end{document}